\documentclass[10pt,twocolumn,letterpaper]{article}

\usepackage{cvpr}
\usepackage{times}
\usepackage{epsfig}
\usepackage{graphicx}
\usepackage{amsmath}
\usepackage{amssymb}
\usepackage{bm}
\usepackage{algorithm}
\usepackage{algorithmic}
\usepackage{verbatim}
\usepackage{multirow}
\usepackage{hhline}

\newcommand{\tabincell}[2]{\begin{tabular}{@{}#1@{}}#2\end{tabular}}

\usepackage[pagebackref=true,breaklinks=true,letterpaper=true,colorlinks,bookmarks=false]{hyperref}

\cvprfinalcopy 

\def\cvprPaperID{8882} 

\ifcvprfinal\fi
\begin{document}

\title{DSNAS: Direct Neural Architecture Search without Parameter Retraining}


\author{
Shoukang Hu*$^{1}$, Sirui Xie*$^{2}$, Hehui Zheng$^{3}$, Chunxiao Liu$^{4}$, Jianping Shi$^{4}$, Xunying Liu$^{1}$, Dahua Lin$^{1}$ \thanks{Equal contribution. Work done at SenseTime Research. Correspondence {\tt skhu@se.cuhk.edu.hk}, {\tt srxie@ucla.edu}}%
\thanks{1. The Chinese Univsersity of Hong Kong; 2. University of California, Los Angeles; 3. Cambridge University; 4. SenseTime Research.}%
}


\maketitle
\thispagestyle{empty}

\begin{abstract}
   If NAS methods are solutions, what is the problem? Most existing NAS methods require two-stage parameter optimization. However, performance of the same architecture in the two stages correlates poorly. In this work, we propose a new problem definition for NAS, task-specific end-to-end, based on this observation. We argue that given a computer vision task for which a NAS method is expected, this definition can reduce the vaguely-defined NAS evaluation to i) accuracy of this task and ii) the total computation consumed to finally obtain a model with satisfying accuracy. Seeing that most existing methods do not solve this problem directly, we propose DSNAS, an efficient differentiable NAS framework that simultaneously optimizes architecture and parameters with a low-biased Monte Carlo estimate. Child networks derived from DSNAS can be deployed directly without parameter retraining. Comparing with two-stage methods, DSNAS successfully discovers networks with comparable accuracy (74.4\%) on ImageNet in 420 GPU hours, reducing the total time by more than 34\%. 
\end{abstract}

\section{Introduction}
The success of deep learning is partially built upon the architecture of neural networks. However, the variation of network architectures always incurs unpredictable changes in performance, causing tremendous efforts in \textit{ad hoc} architecture design. Neural Architecture Search (NAS) is believed to be promising in alleviating this pain. Practitioners from the industry would like to see NAS techniques that automatically discover task-specific networks with reasonable performance, regardless of their generalization capability. Therefore, NAS is always formulated as a hyper-parameter optimization problem, whose algorithmic realization spans evolution algorithm \cite{stanley2002evolving, guo2019single}, reinforcement learning \cite{zoph2016neural}, Bayesian optimization \cite{kandasamy2018neural}, Monte Carlo Tree Search \cite{wistuba2017finding}, and differentiable architecture search \cite{liu2018darts, xie2018snas, cai2018proxylessnas}. Recently, these algorithmic frameworks have exhibited pragmatic success in various challenging tasks, e.g. semantic segmentation \cite{liu2019auto} and object detection \cite{chen2019detnas} \etc.  

\begin{figure}[!t]
    \centering
    \includegraphics[width=3in]{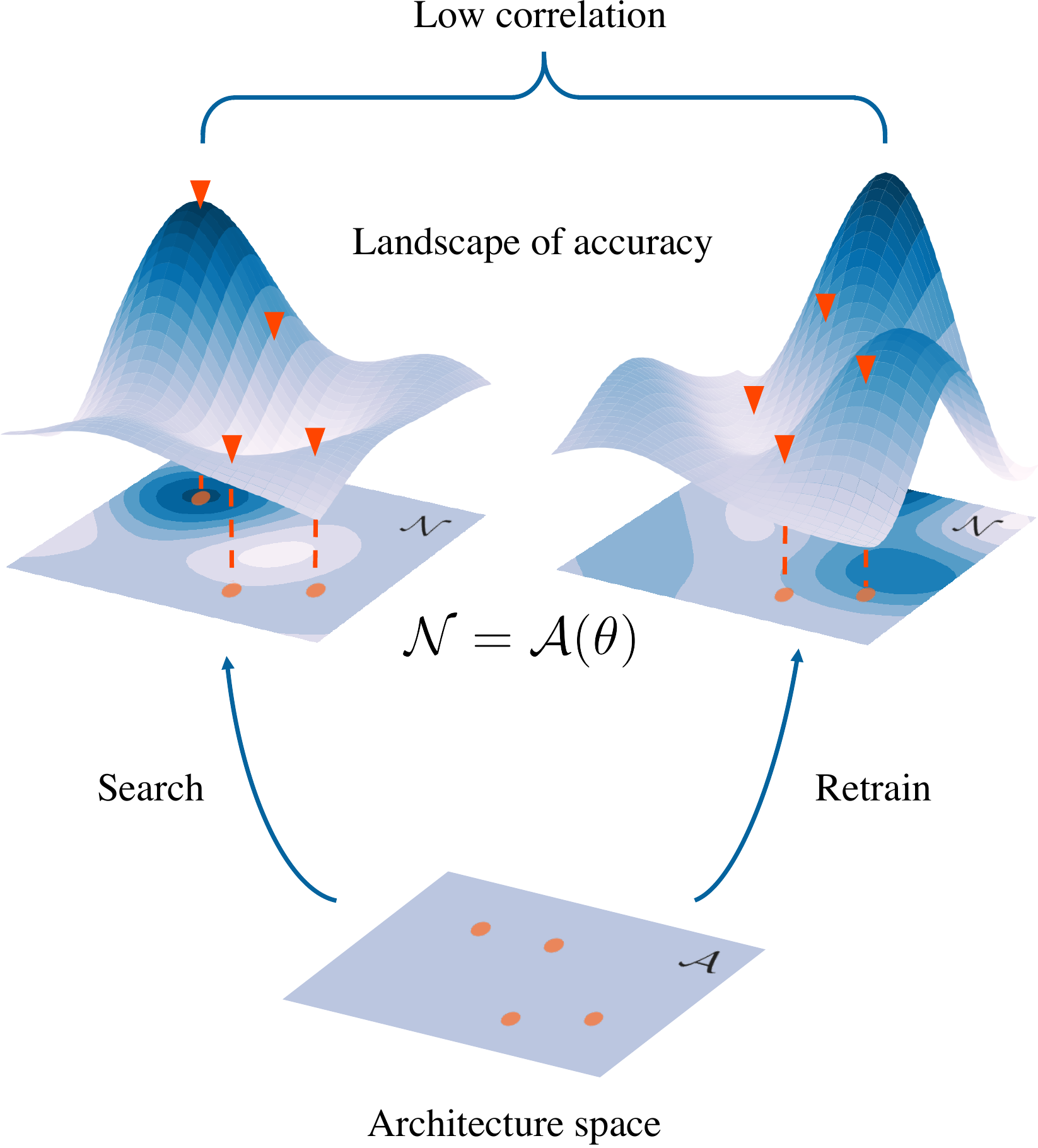}
    \caption{Projecting from the architecture space $\mathcal{A}$ to the network space $\mathcal{N}(\theta)$ with different parameter training schemes in \textit{searching} and \textit{retraining} results in accuracy with low correlation.}
    \label{fig:correlation}
\end{figure}

However, even as an optimization problem, NAS is almost vaguely defined. Most of the NAS methods proposed recently are implicitly two-stage methods. These two stages are \textit{searching} and \textit{evaluation} (or \textit{retraining}). While the architecture optimization process is referring to the \textit{searching} stage, in which a co-optimization scheme is designed for parameters and architectures, there runs another round of parameter optimization in the \textit{evaluation} stage, on the same set of training data for the same task. This is to some extent contradicting the norm in a machine learning task that no optimization is allowed in \textit{evaluation}. A seemingly sensible argument could be that the \textit{optimization} result of NAS is only the architecture, and the \textit{evaluation} of an architecture is to check its performance after retraining. There is certainly no doubt that architectures that achieve high performance when retrained from scratch are reasonable choices for deployment. But 
is this search method still valid if the searched architecture does not perform well after retraining, due to the inevitable difference of training setup in \textit{searching} and \textit{evaluation}? 

These questions can only be answered with an assumption that the final \textit{searching} performance can be generalized to \textit{evaluation} stage even though the training schemes in two stages are different. Specifically, differences in training schemes may include different number of cells, different batch sizes, and different epoch numbers, \etc. Using parameter sharing with efficiency concerns during search is also a usual cause. Unfortunately, this assumption is not a valid one. The correlation between the performance at the end of \textit{searching} and after the retraining in \textit{evaluation} is fairly low, as long as the parameter-sharing technique is used \cite{sciuto2019evaluating, chu2019fairnas}. 

We are thus motivated to rethink the problem definition of neural architecture search. We want to argue that as an application-driven field, there can be a diverse set of problem definitions, but every one of them should not be vague. And in this work, we put our cards on the table: we aim to tackle the \textit{task-specific end-to-end} NAS problem. Given a task, defined by a data set and an objective (\eg training loss), the expected NAS solution optimizes architecture and parameters to automatically discover a neural network with reasonable (if not optimal by principle) performance. By the term \textit{end-to-end}, we highlight the solution only need a single-stage training to obtain a \textit{ready-to-deploy} neural network of the given task. And the term \textit{task-specific} highlights the boundary of this solution. The searched neural network can only handle this specific task. We are not confident whether this neural network generalizes well in other tasks. Rather, what can be expected to generalize is this NAS framework.  

Under this definition, the evaluation metrics of a proposed framework become clear, namely searching efficiency and final performance. Scrutinizing most existing methods in these two metrics, we find a big niche for a brand new framework. On one side of the spectrum, gradient-based methods such as ENAS \cite{pham2018efficient}, DARTS \cite{liu2018darts}, ProxylessNAS \cite{cai2018proxylessnas} require two-stage parameter optimization. This is because in the approximation to make them differentiable, unbounded bias or variance are introduced to their gradients. Two-stage methods always consume more computation than single-stage ones, not only because of another round of training but also the reproducibility issue \cite{li2019random}. On the other side of the spectrum, one-shot methods such as random search \cite{li2019random} and SPOS \cite{guo2019single} can be extended to single-stage training. But since they do not optimize the architecture distribution in parameter training, the choice of prior distribution becomes crucial. A uniform sampling strategy may potentially subsume too many resources for satisfying accuracy. Lying in the middle, SNAS \cite{xie2018snas} shows a proof of concept, where the derived network maintains the performance in the \textit{searching} stage. However, the gumbel-softmax relaxation makes it necessary to store the whole parent network in memory in both forward and backward, inducing tremendous memory and computation waste.   

In this work, we confront the challenge of single-stage simultaneous optimization on architecture and parameters. Our proposal is an efficient differentiable NAS framework, Discrete Stochastic Neural Architecture Search (DSNAS). Once the search process finishes, the best-performing subnetwork is derived with optimized parameters, and no further retraining is needed. DSNAS is built upon a novel search gradient, combining the stability and robustness of differentiable NAS and the memory efficiency of discrete-sampling NAS. This search gradient is shown to be equivalent to SNAS's gradient at the discrete limit, optimizing the \textit{task-specific end-to-end} objective with little bias. And it can be calculated in the same round of back-propagation as gradients to neural parameters. Its forward pass and back-propagation only involve the compact subnetwork, whose computational complexity can be shown to be much more friendly than DARTS, SNAS and even ProxylessNAS, enabling large-scale direct search. We instantiate this framework in a single-path setting. The experimental results show that DSNAS discovers networks with comparable performance ($74.4\%$) in ImageNet classification task in only \bm{$420$} GPU hours, reducing the total time of obtaining a \textit{ready-to-deploy} solution by $34\%$ from two-stage NAS.

 To summarize, our main contributions are as follows:
\begin{itemize}
    \item We propose a well-defined neural architecture search problem, \textit{task-specific end-to-end} NAS, under the evaluation metrics of which most existing NAS methods still have room for improvement. 
    \item We propose a \textit{plug-and-play} NAS framework, DSNAS, as an efficient solution to this problem in large scale. DSNAS updates architecture parameters with a novel search gradient, combining the advantages of policy gradient and SNAS gradient. A simple but smart implementation is also introduced. 
    \item We instantiate it in a single-path parent network. The empirical study shows DSNAS robustly discovers neural networks with state-of-the-art performance in ImageNet, reducing the computational resources by a big margin over two-stage NAS methods. We have made our implementation public\footnote{https://github.com/SNAS-Series/SNAS-Series/}.  
\end{itemize}

\section{Problem definition of NAS}

\subsection{Two-Stage NAS}
Most existing NAS methods involve optimization in both \textit{searching} stage and \textit{evaluation} stage. In the \textit{searching} stage, there must be parameter training and architecture optimization, even though they may not run simultaneously. The ideal way is to train all possible architectures from scratch and then select the optimal one. However, it is infeasible with the combinatorial complexity of architecture. Therefore, designing the co-occurrence of parameter and architecture optimization to improve efficiency is the main challenge of any general NAS problems. This challenge has not been overcome elegantly yet. The accuracy at the end of the \textit{searching} stage has barely been reported to be satisfying. And an \textit{ad hoc} solution is to perform another round of parameter optimization in the \textit{evaluation} stage. 

Optimizing parameters in \textit{evaluation} stage is not normal in traditional machine learning. Normally, the data set provided is divided into training set and validation set. Ones do learning in the \textit{training} stage, with data from the training set. Then the learned model is tested on the withheld validation set, where no further training is conducted. With the assumption that training data and validation data are from the same distribution, the learning problem is reduced to an optimization problem. Ones can hence be confident to expect models with high training accuracy, if the assumption is correct, have high evaluation accuracy. 

Allowing parameter retraining in the \textit{evaluation} stage makes NAS a vaguely defined machine learning problem. Terming problems as \textit{Neural Architecture Search} give people an inclined interpretation that only the architecture is the learning result, instead of parameters. But if the searched architecture is the answer, what is the problem? Most NAS methods claim they are discovering \textit{best-performing} architecture in the designated space efficiently \cite{cai2018proxylessnas, guo2019single, kandasamy2018neural}, but what specifically does \textit{best-performing} mean? Given that retraining is conducted in evaluation stage, ones may naturally presume it is a meta-learning-like hyperparameter problem. Then the optimization result should exhibit some meta-level advantages, such as faster convergence, better optimum or higher transferability, etc. These are objectives that ones are supposed to state clearly in a NAS proposal. Nonetheless, objectives are only implicitly conveyed (mostly better optimum) in experiments.

Defining problem precisely is one of the milestones in scientific research, whose direct gift in a machine learning task is a clear objective and evaluation metric. Subsequent efforts can then be devoted into validating if the proposed learning loss is approximating a \textit{necessary and sufficient} equivalence of this objective. Unfortunately, under this criterion, most existing two-stage NAS methods are reported \cite{sciuto2019evaluating, li2019random} failing to prove the correlation between the \textit{searching} accuracy and the \textit{retraining} accuracy.

\subsection{Task-specific end-to-end NAS}
Seeing that the aforementioned dilemma lies in the ambiguity in evaluating an architecture alone, we propose a type of problem termed \textit{task-specific end-to-end NAS}, the solution to which should provide a \textit{ready-to-deploy} network with optimized architecture and parameters.

\textbf{Task} refers to generally any machine learning tasks (in this work we discuss computer vision tasks specifically). A well-defined task should at least have a set of data representing its functioning domain, a learning objective for the task-specific motives \textit{e.g.} classification, segmentation, etc. And the task is overwritten if there is a modification in either factor, even a trivial augmentation in the data. In other words, \textit{task-specific} sets a boundary on what we can expect from the searched result and what cannot. This can bring tremendous operational benefits to industrial applications.  

\textbf{End-to-end} highlights that, given a task, the expected solution can provide a \textit{ready-to-deploy} network with satisfying accuracy, the whole process of which can be regarded as a black-box module. Theoretically, it requires a direct confrontation of the main challenge of any general NAS problem, \textit{i.e.} co-optimizing parameter and architecture efficiently. Empirically, \textit{task-specific end-to-end} is the best description of NAS's industrial application scenarios: i) the NAS method itself should be generalizable for any off-the-shelf tasks; and ii) when applied to a specific task, practitioners can at least have some conventional guarantees on the results. Basically, it is to reduce vaguely defined NAS problems to established tasks. 

The evaluation metrics become clear under this problem definition. The performance of the final result is, by principle, the accuracy in this task. And the efficiency should be calculated based on the time from this NAS solver starts taking data to it outputs the neural network whose architecture and parameters are optimized. This efficiency metric is different from all existing works. For two-stage methods, the time for both \textit{searching} and \textit{evaluation} should be taken into account in this metric. Therefore, their efficiency may not be as what they claim. Moreover, two-stage methods do not optimize the objective \textit{higher accuracy of final derived networks} in an end-to-end manner.  

\section{Direct NAS without retraining}

\subsection{Stochastic Neural Architecture Search (SNAS)}

In the literature, SNAS is one of those close to a solution to the \textit{task-specific end-to-end NAS} problem. Given any task with differentiable loss, the SNAS framework directly optimizes the expected performance over architectures in terms of this task. In this subsection, we provide a brief introduction on SNAS.

\begin{figure*}[tp]
    \centering
    \includegraphics[width=0.95\textwidth]{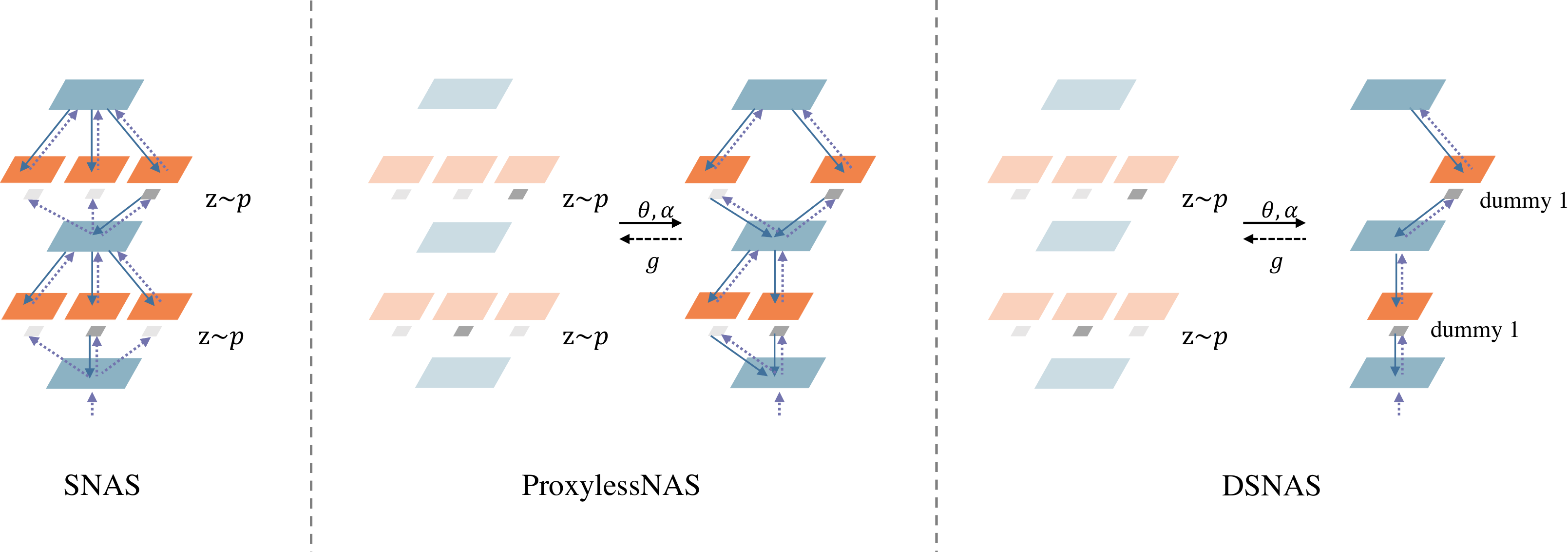}
    \caption{Forward and backward on SNAS, ProxylessNAS and DSNAS. Blue lumps stand for feature maps, orange ones for operation candidates. Blue arrow lines indicate forward data flows, purple dashed lines indicate backward ones. Semi-transparent lumps stand for parent networks that are not instantiated with \textit{batch} dimension, a technique to reduce computation in ProxylessNAS and DSNAS. $dummy$ $1$ highlights the smart implementation introduced in Sec. 3.3. }
    \label{fig:disc_snas}
\end{figure*}

Basically, SNAS is a differentiable NAS framework that maintains the generative nature as reinforcement-learning-based methods \cite{zoph2016neural}. Exploiting the deterministic nature of the Markov Decision Process (MDP) of network construction process, SNAS reformulated it as a Markov Chain. This reformulation leads to a novel representation of the network construction process. As shown in Fig.\ref{fig:disc_snas}, nodes $x_{i}$ (blue lumps) in the DAG represent feature maps. Edges $(i, j)$ (arrow lines) represent information flows between nodes $x_{i}$ and $x_{j}$, on which $n$ possible operations $\bm{O}_{i, j}$ (orange lumps) are attached. Different from DARTS, which avoids sampling subnetwork with an attention mechanism, SNAS instantiates this Directed Acyclic Graph (DAG) with a stochastic computational graph. Forwarding a SNAS parent network is to first sample random variables $\bm{Z}_{i, j}$ and multiplying it to edges $(i, j)$ in the DAG:
\begin{equation}
\tilde{\bm{O}}_{i, j}(\cdot)= \bm{Z}_{i,j}^{T}\bm{O}_{i, j}(\cdot).
\end{equation}
Ones can thus obtain a Monte Carlo estimate of the expectation of task objective $L_{\bm{\theta}}(\bm{Z})$ over possible architectures:
\begin{equation}
 \mathbb{E}_{\bm{Z}\sim p_{\bm{\alpha}}(\bm{Z})}[L_{\bm{\theta}}(\bm{Z})],
\label{eq:objective}
\end{equation}
where $\bm{\alpha}$ and $\bm{\theta}$ are parameters of architecture distribution and neural operations respectively. This is exactly the \textit{task-specific end-to-end NAS} objective. 

To optimize parameters $\bm{\theta}$ and architecture $\bm{\alpha}$ simultaneously with Eq. \ref{eq:objective}, (termed as \textit{single-level optimization} in \cite{liu2018darts}), SNAS relaxes the discrete one-hot random variable ${\bm{Z}}$ to a continuous random variable $\tilde{\bm{Z}}$ with the gumbel-softmax trick. However, the continuous relaxation requires to store the whole parent network in GPU, preventing it from directly applying to large-scale networks. In Xie et al. \cite{xie2018snas}, SNAS is still a two-stage method.

If the temperature in SNAS's gumble-softmax trick can be directly pushed to zero, SNAS can be extended to large-scale networks trivially. However, it is not the case. Take a look at the search gradient given in Xie et al. \cite{xie2018snas}:
\begin{equation}
\frac{\partial \mathcal{L}}{\partial \alpha_{i, j}^{k}} = \frac{\partial \mathcal{L}}{\partial x_{j}}\bm{O}_{i, j}^{T}(x_{i}) (\bm{\delta}(k'-k)-\tilde{\bm{Z}}_{i, j})Z_{i, j}^{k}\frac{1}{\lambda \alpha_{i, j}^{k}},
\label{eq:snas_search_grad}
\end{equation}
ones can see that the temperature $\lambda$ is not valid to be zero for the search gradient. Xie et al. \cite{xie2018snas} only gradually annealed it to be close to zero. In this work, we seek for an alternative way to differentiate Eq. \ref{eq:objective}, combining the efficiency of discrete sampling and the robustness of continuous differentiation. And we start from SNAS's credit assignment. 

\subsection{Discrete SNAS (DSNAS)}
In original SNAS \cite{xie2018snas}, to prove its efficiency over ENAS, a policy gradient equivalent of the search gradient is provided
\begin{equation}
\begin{split}
&\mathbb{E}_{\tilde{\bm{Z}}\sim p(\tilde{\bm{Z}})}[\frac{\partial \mathcal{L}}{\partial \alpha_{i,j}^{k}}]\\
&= \mathbb{E}_{\tilde{\bm{Z}}\sim p(\tilde{\bm{Z}})}[\nabla_{\alpha_{i,j}^{k}}\log p(\tilde{\bm{Z}})[\frac{\partial \mathcal{L}}{\partial x_{j}}\bm{O}_{i,j}^{T}(x_{i}) \tilde{\bm{Z}}_{i, j}]_{c}],
\label{eq:cost_obj_cont_snas}
\end{split}
\end{equation}
where $\tilde{\bm{Z}}_{i,j}$ is the gumbel-softmax random variable, $[\cdot]_{c}$ denotes that $\cdot$ is a \textit{cost} independent from $\bm{\alpha}$ for gradient calculation. In other words, Eq. \ref{eq:cost_obj_cont_snas} and Eq. \ref{eq:snas_search_grad} both optimize the \textit{task-specific end-to-end NAS} objective \textit{i.e.} Eq. \ref{eq:objective}. 


In order to get rid of SNAS's continuous relaxation, we push the $\lambda$ in the PG equivalent (\ref{eq:cost_obj_cont_snas}) to the limit $0$, with the insight that only reparameterization trick needs continuous relaxation but policy gradient doesn't. The expected \textit{search gradient} for architecture parameters at each edge becomes:
\begin{equation}
\begin{split}
&\lim_{\lambda \to 0}\mathbb{E}_{\tilde{\bm{Z}}\sim p(\tilde{\bm{Z}})}[\frac{\partial \mathcal{L}}{\partial \alpha_{i,j}^{k}}]\\
&= \lim_{\lambda \to 0}\mathbb{E}_{\tilde{\bm{Z}}\sim p(\tilde{\bm{Z}})}[\nabla_{\alpha_{i,j}^{k}}\log p(\tilde{\bm{Z}}_{i,j})[\frac{\partial \mathcal{L}}{\partial x_{j}}\tilde{\bm{O}}_{i, j}(x_{i})]_{c}]\\
&= \mathbb{E}_{\bm{Z}\sim p(\bm{Z})}[\nabla_{\alpha_{i,j}^{k}}\log p(\bm{Z}_{i,j})[\frac{\partial L}{\partial x_{j}}\bm{O}^{T}_{i, j}(x_{i})\bm{Z}_{i,j}]_{c}]\\
&= \mathbb{E}_{\bm{Z}\sim p(\bm{Z})}[\nabla_{\alpha_{i,j}^{k}}\log p(\bm{Z}_{i,j})[\frac{\partial L}{\partial x_{j}}\sum_{k}{O}^{k}_{i, j}(x_{i}){Z}^{k}_{i,j}]_{c}],
\label{eq:snas_disc}
\end{split}
\end{equation}
where $\bm{Z}_{i,j}$ is a strictly one-hot random variable, ${Z}^{k}_{i,j}$ is the $k$th element in it, $[\cdot]_{c}$ denotes that $\cdot$ is a \textit{cost} independent from $\bm{\alpha}$ for gradient calculation. Line 3 is derived from line 2 since $p(\bm{Z}_{i,j})=\lim_{\lambda \to 0}p(\tilde{\bm{Z}}_{i,j})$ \cite{maddison2016concrete}, $L = \lim_{\lambda \to 0}\mathcal{L}$.



Exploiting the one-hot nature of $\bm{Z}_{i,j}$, \textit{i.e.} only $Z^{s}_{i,j}$ on edge $(i, j)$ is 1, others \textit{i.e.} $Z^{\smallsetminus s}_{i,j}$ are $0$, the \textit{cost} function can be further reduced to
\begin{equation}
\begin{split}
C(\bm{Z}_{i,j}) &=\sum_{k}\frac{\partial L}{\partial x_{j}}{O}^{k}_{i, j}(x_{i}){Z}^{k}_{i,j}\\
&= \frac{\partial L}{\partial x_{j}^{i}}O^{s}_{i, j}(x_{i}){Z}^{s}_{i,j} = \frac{\partial L}{\partial x_{j}^{i}}x_{j}^{i} \\
&= \frac{\partial L}{\partial x_{j}^{i}}\frac{\partial x_{j}^{i}}{\partial Z^{s}_{i,j}} = \frac{\partial L}{\partial Z^{s}_{i,j}}, 
\end{split}
\label{eq:cost_disc}
\end{equation}
as long as $|\frac{\partial \mathcal{L}}{\partial x_{j}}{O}^{\smallsetminus s}_{i, j}(x_{i})| \ne \infty$. Here $x_{j}^{i}=O^{s}_{i, j}(x_{i})Z^{s}_{i,j}$ is the output of the operation $O^{s}_{i, j}$ chosen at edge $(i, j)$. The equality in line 3 is due to $Z^{s}_{i,j}=1$.

\subsection{Implementation}

The algorithmic fruit of the mathematical manipulation in Eq. \ref{eq:cost_disc} is a parallel-friendly implementation of Discrete SNAS, as illustrated in Fig. \ref{fig:disc_snas}. In SNAS, the network construction process is a pass of forward of stochastic computational graph. The whole network has to be instantiated with the \textit{batch} dimension. In DSNAS we offer an alternative implementation. Note that $C(\bm{Z}_{i,j})=\frac{\partial L}{\partial Z^{s}_{i,j}}$ only needs to be calculated for the sampled subnetworks. And apparently it is also the case for $\frac{\partial L}{\partial \bm{\theta}}$. That is to say, the back-propagation of DSNAS only involves the sampled network, instead of the whole parent network. Thus we only instantiate the subnetwork with the \textit{batch} dimension for forward and backward. However, the subnetwork derived in this way does not necessarily contain $\bm{Z}_{i,j}$. If it was not with Line 3 of Eq. \ref{eq:cost_disc}, we would have to calculate $C(\bm{Z}_{i,j})$ with $\frac{\partial L}{\partial x_{j}^{i}}x_{j}^{i}$. Then the policy gradient loss would explicitly depend on the intermediate result $x_{j}^{i}=O^{s}_{i, j}(x_{i})$, which may need an extra round of forward if it is not stored by the automated differentiation infrastructure. With a smart mathematical manipulation in Eq. \ref{eq:cost_disc}, ones can simply multiply a $dummy$ $1$ to the output of each selected operation, and calculate $C(\bm{Z}_{i,j})$ with $\frac{\partial L}{\partial 1^{dummy}_{i,j}}$. The whole algorithm is shown in Alg. \ref{alg:dis_snas} 
    \begin{algorithm}[t]\small
    \caption{Discrete SNAS}
    \label{alg:dis_snas}
    \begin{algorithmic} 
    \REQUIRE parent network, operation parameters $\bm{\theta}$ and categorical arch distribution $p_{\bm{\alpha}}(\bm{Z})$
    \STATE Initialize $\bm{\theta}$, $\bm{\alpha}$ 
    \WHILE{not converged} 
    \STATE Sample one-hot random variables $\bm{Z}$ from $p_{\bm{\alpha}}(\bm{Z})$
    \STATE Construct child network with $\bm{\theta}$ according to $\bm{Z}$, multiply a $1^{dummy}$ after each feature map $X$ 
    \STATE Get a batch from data and forward to get $L$
    \STATE Backward $L$ to both $\bm{\theta}$ and $1^{dummy}$, backward $\log p_{\bm{\alpha}}(\bm{Z})$ to $\bm{\alpha}$
    \STATE Update $\bm{\theta}$ with $\frac{\partial L}{\partial \bm{\theta}}$, update $\bm{\alpha}$ with $\frac{\partial \log p_{\bm{\alpha}}(\bm{Z})} {\partial \bm{\alpha}}  \frac{\partial L}{\partial 1^{dummy}}$
    \ENDWHILE
    \end{algorithmic}
    \end{algorithm}

\subsection{Complexity analysis}
In this subsection, we provide a complexity analysis of DSNAS, SNAS, and ProxylessNAS.  Without loss of generality, we define a parent network with $l$ layers and each layer has $n$ candidate choice blocks. Let the forward time on a sampled subnetwork be $P$, its backward time be $Q$, and the memory requirement for this round be $M$. 

As the original SNAS instantiates the whole graph with \textit{batch} dimension, it needs $n$ times GPU memory and $n$ times calculation comparing to a subnetwork. It is the same case in DARTS.   

\begin{table}[h!]
\centering
\begin{tabular}{c|c|c|c}
\hline
Method  & \tabincell{c}{Forward\\ time} & \tabincell{c}{Backward\\ time} & \tabincell{c}{Memory} \\
\hline\hline
Subnetwork & $O(P)$ & $O(Q)$ & $O(M)$\\
SNAS & $O(nP)$ & $O(nQ)$ & $O(nM)$ \\
ProxylessNAS* & $O(nP)$ & $O(nQ)$ & $O(nM)$ \\
ProxylessNAS  & $O(2P)$ & $O(2Q)$ & $O(2M)$\\
DSNAS  & $O(P)$ & $O(Q)$ & $O(M)$\\
\hline
\end{tabular}
\caption{Computation complexity comparison between SNAS, ProxylessNAS and DSNAS. ProxylessNAS* indicates its theoretical complexity.}\label{tab:complexity}
\end{table}

This memory consumption problem of differentiable NAS was first raised by \cite{cai2018proxylessnas}. And they proposed an approximation to DARTS's optimization objective, with the BinaryConnect \cite{courbariaux2015binaryconnect} technique: 
\begin{equation}
    \frac{\partial \mathcal{L}}{\partial \bm{\alpha}_{i,j}^{}} = \frac{\partial \mathcal{L}}{\partial \hat{\bm{Z}}_{i,j}}\frac{\partial \hat{\bm{Z}}_{i,j}}{\partial \bm{\alpha}_{i,j}^{}} \approx \sum_{k}\frac{\partial \mathcal{L}}{\partial {{Z}}_{i,j}^{k}}\frac{\partial \hat{{Z}}_{i,j}^{k}}{\partial \bm{\alpha}_{i,j}^{}},
\label{eq:proxyless}
\end{equation}
where $\hat{\bm{Z}}_{i,j}$ denotes the attention-based estimator as in DARTS \cite{liu2018darts}, distinct from the discrete random variable ${\bm{Z}}_{i,j}$, highlighting how the approximation is being done. But this approximation does not directly save the memory and computation. Different from Eq. \ref{eq:snas_disc} and Eq. \ref{eq:cost_disc}, theoretically, the calculation of Eq. \ref{eq:proxyless} still involves the whole network, as indicated by the summation $\sum$. To reduce memory consumption, they further empirically proposed a path sampling heuristic to decrease the number of paths from $n$ to $2$. Table \ref{tab:complexity} shows the comparison.

\subsection{Progressive early stop}
One potential problem in sample-based differentiable NAS is that empirically, the entropy of architecture distribution does not converge to zero, even though comparing to attention-based NAS \cite{liu2018darts} they are reported \cite{xie2018snas} to converge with smaller entropy. The non-zero entropy keeps the sampling going on until the end, regardless of the fact that sampling at that uncertainty level does not bring significant gains. To the opposite, it may even hinder the learning on other edges. 

To avoid this side-effect of architecture sampling, DSNAS applies a progressive early stop strategy. Sampling and optimization stop at layers/edges in a progressive manner. Specifically, a threshold $h$ is set for the stopping condition:
\begin{equation}
    \begin{aligned}
    \label{eq:threshold}
     \min\{\alpha_{i,j}^{k} - \alpha_{i,j}^{m}, \forall m \neq k || \alpha_{i,j}^{k}=\max\{\bm{\alpha}_{i,j}\} \} \geq h.
    \end{aligned}
\end{equation}
Once this condition is satisfied on any edge/layer, we directly select the operation choice with the highest probability there, stop its sampling and architecture parameters update in the following training.

\subsection{Comparison with one-shot NAS}
Different from all differentiable NAS methods, one-shot NAS only do architecture optimization in one-shot, before which they obtain a rough estimation of the graph through either pretraining \cite{bender2018understanding, guo2019single} or an auxiliary hypernetwork \cite{brock2017smash}. All of them are two-stage methods. The advantage of DSNAS is that it optimizes architecture alongside with parameters, which is expected to save some resources in the pretraining stage. Intuitively, DSNAS rules out non-promising architectures in an adaptive manner by directly optimizing the objective in an end-to-end manner. Although one-shot methods can also have an end-to-end realization, by investing more resources in pretraining, it may take them more epochs to achieve comparable performance as DSNAS. They can also do finetuning, but still parameters of the optimal networks are updated less frequently than DSNAS. Ones can expect better performance from DSNAS given equivalent training epochs.

\section{Experimental Results}
In this section, we first demonstrate why the proposed \textit{task-specific end-to-end} is an open problem for NAS, by investigating the performance correlation between \textit{searching} stage and \textit{evaluation} stage of the two-stage NAS. We then validate the effectiveness and efficiency of DSNAS under the proposed \textit{task-specific end-to-end} metric on the same search space as SPOS \cite{guo2019single}. We further provide a breakup of time consumption to illustrate the computational efficiency of DSNAS.  

\subsection{Accuracy correlation of two-stage NAS}
Since the validity of the \textit{searching} in two-stage NAS relies on a high correlation in the performance of \textit{searching} stage and \textit{evaluation} stage, we check this assumption with a ranking correlation measure, Kendall Tau metric $\tau$ \cite{kendall1938new}.
\begin{equation}
    \begin{aligned}
    \label{eq:Kendall_Tau}
     \tau=\frac{2(N_{concordant}-N_{disconcordant})}{N(N-1)},
    \end{aligned}
\end{equation}
where $N$ is the total number of pairs ($x_{i}, y_{i}$) from the \textit{searching} stage and \textit{evaluation} stage consisting of $N_{concordant}$ concordant ranking pairs ($x_1 > x_2, y_1 > y_2$ or $x_1 < x_2, y_1 < y_2$) and $N_{disconcordant}$ disconcordant ranking pairs ($x_1 > x_2, y_1 < y_2$ or $x_1 < x_2, y_1 > y_2$). Kendall Tau metric ranges from -1 to 1, which means the ranking order changes from reversed to identical. $\tau$ being close to 0 indicates the absence of correlation. 

We measure the ranking correlation by calculating Kendall Tau metric from two perspectives: (1) The $\tau_{inter}$ is calculated based on the top-k model performance of the \textit{searching} and \textit{evaluation} stage in one single searching process; (2) The Kendal Tau metric $\tau_{intra}$ is calculated by running the two-stage NAS methods several times with different random seeds using the top-1 model performance in each searching process. As shown in Table \ref{Kendall_Tau_metric}, the performance correlation between the \textit{searching} stage and \textit{evaluation} stage in both SPOS and ProxylessNAS is fairly low. This indicates the necessity of \textit{task-specific end-to-end} NAS problem formulation. Fairly low correlation may also imply reproducibility problems.
\begin{table}[h!]
\centering
\begin{tabular}{c|c|c}
\hline
Model & $\tau_{inter}$ & $\tau_{intra}$ \\ 
\hline\hline
Single Path One-Shot\cite{guo2019single} & 0.33  &  -0.07\\
ProxylessNas \cite{cai2018proxylessnas} & - & -0.33 \\
\hline
\end{tabular}
\caption{Kendall Tau metric $\tau$ calculated with the top-k model performance in the searching and evaluation stage. $\tau_{inter}$ measures the correlation of top-k model performance of the \textit{searching} and \textit{evaluation} stage in one single searching process while $\tau_{intra}$ measures the correlation of top-1 model performance from different searching processes.}\label{Kendall_Tau_metric}
\end{table}

\begin{figure*}[t!]
    \centering
    \includegraphics[width=3.3in]{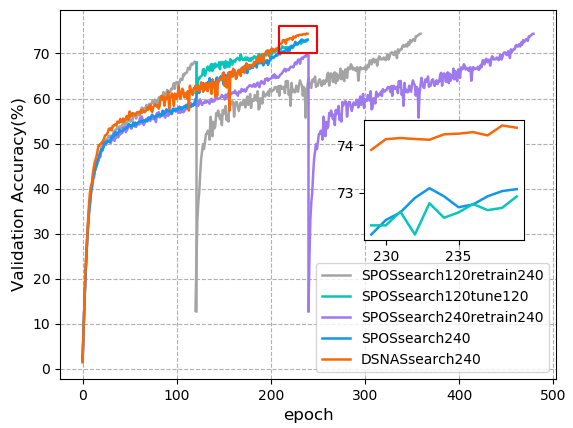}
    \includegraphics[width=3.3in]{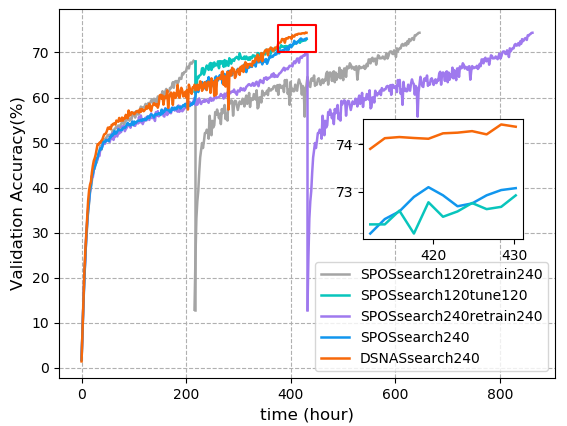}
    \caption{Searching process of two-stage SPOS and single-stage SPOS/DSNAS. \texttt{SPOSsearch120retrain240} and \texttt{SPOSsearch240retrain240} search for 120/240 epochs then retrain the derived architecture for 240 epochs.
    Instead of retraining, \texttt{SPOSsearch120tune120} finetunes the result for 120 epochs.  \texttt{DSNASsearch240} and \texttt{SPOSsearch240} utilize one-stage training for 240 epochs. \texttt{DSNASsearch240} applies progressive early stop, \texttt{SPOSsearch240} applies one-shot EA at 120th epoch. }
    \label{fig:search_train_comparison_figure}
\end{figure*}

\subsection{Single-path architecture search}
\noindent
\textbf{Motivation  } 
To compare the efficiency and accuracy of derived networks from DSNAS versus existing two-stage methods, we conduct experiment in single-path setting. Results are compared in the \textit{task-specific end-to-end} metrics. 
 
\noindent
\textbf{Dataset  } 
All our experiments are conducted in a mobile setting on the ImageNet Classification task \cite{russakovsky2015imagenet} with a resource constraint $FLOPS \leq 330M$. This dataset consists of around $1.28\times10^6$ training images and $5\times10^4$ validation images. Data transformation is achieved by the standard pre-processing techniques described in the supplementary material.

\noindent
\textbf{Search Space  } 
The basic building block design is inspired by ShuffleNet v2 \cite{ma2018shufflenet}. There are 4 candidates for each choice block in the parent network, i.e., choice\_3, choice\_5, choice\_7, and choice\_x. These candidates differ in the kernel size and the number of depthwise convolutions, spanning a search space with $4^{20}$ single path models. The overall architecture of the parent network and building blocks are shown in the supplementary material. 

\noindent
\textbf{Training Settings  } 
We follow the same setting as SPOS \cite{guo2019single} except that we do not have an \textit{evaluation stage} in our searching process. We adopt a SGD optimizer with a momentum of 0.9 \cite{sutskever2013importance} to update the parent network weight parameters. A cosine learning rate scheduler with an initial learning rate of 0.5 is applied. Moreover, an L2 weight decay ($4\times10e^{-5}$) is used to regularize the training process. The architecture parameters are updated using the Adam optimizer with an initial learning rate of 0.001. All our experiments are done on 8 NVIDIA TITAN X GPUs. 

\noindent
\textbf{Searching Process  } 
To demonstrate the efficiency of DSNAS, we compare the whole process needed to accomplish \textit{task-specific end-to-end} NAS in ImageNet classification with two-stage NAS methods. Among all existing two-stage methods, SPOS \cite{guo2019single} is the one with state-of-the-art accuracy and efficiency. SPOS can be regarded as a weight-sharing version of random search, where the search is conducted with one-shot evolution algorithm after training the uniformly sampled parent network.  

Figure \ref{fig:search_train_comparison_figure} shows DSNAS's advantage over several different configurations of SPOS. We purposefully present curves in terms of both epoch number and time to illustrate that even though DSNAS updates architecture in an iteration-basis, almost no extra computation time is introduced. Among the four configurations of SPOS, \texttt{SPOSsearch120retrain240} is the original one as in Guo et al. \cite{guo2019single}, using the two-stage paradigm. Obviously, DSNAS achieves comparable accuracy in an end-to-end manner, with roughly $34\%$ less computational resources. As \texttt{SPOSsearch120retrain240} updates block parameters for only 120 epochs\footnote{Same learning rate scheduler is used in DSNAS and SPOS.}, we run the \texttt{SPOSsearch120tune120} and \texttt{SPOSsearch240retrain240} configurations for fair comparison. At the end of the 240th epoch, the accuracy of SPOS models is around $1.4\%$ and $4\%$ lower than DSNAS's respectively.

In addition, for the ablation study of DSNAS's progressive early stop strategy, we call the EA algorithm of SPOS at the 120th epoch in the one-stage \texttt{DSNASsearch240} configuration. Continuing the parameter training, the selected models experience a leap in accuracy and converge with accuracy $1.3\%$ lower than DSNAS's. However, seeking this EA point is fairly \textit{ad hoc} and prone to random noise.

\noindent
\textbf{Searching Results} 
The experimental results are reported in Table \ref{CB_result}. Comparing with all existing two-stage NAS methods, DSNAS shows comparable performance using at least 1/3 less computational resources. More importantly, the standard deviation in DSNAS's accuracy is lower than those from both \textit{searching} and \textit{evaluation} stage from EA-based SPOS (0.22 vs 0.38/0.36). This exhibits as a differentiable NAS framework, DSNAS is a more robust method in the \textit{task-specific end-to-end} metric. 

\begin{table*}[h!]
\centering
\begin{tabular}{c|c|c|c|c|c|c|c|c}
\hline
\multirow{3}*{Model} & \multirow{3}*{FLOPS} & \multicolumn{2}{c|}{Search} & \multicolumn{2}{c|}{Retrain} & \multirow{3}*{\tabincell{c}{No \\ proxy}} & \multicolumn{2}{c}{Time (GPU hour)} \\
\cline{3-6}\cline{8-9}
 & & \tabincell{c}{Top-1 \\ acc(\%)} & \tabincell{c}{Top-5 \\ acc(\%)} & \tabincell{c}{Top-1 \\ acc(\%)} & \tabincell{c}{Top-5 \\ acc(\%)} & & Search & Retrain\\
\hline\hline
MobileNet V1 (0.75x)\cite{howard2017mobilenets} & 325M & \multicolumn{2}{c|}{Manual} &  68.4 & - & \multicolumn{3}{c}{Manual} \\
MobileNet V2 (1.0x)\cite{sandler2018mobilenetv2} & 300M & \multicolumn{2}{c|}{Manual} &  72.0 & 91.00 & \multicolumn{3}{c}{Manual}\\
ShuffleNet V2 (1.5x)\cite{ma2018shufflenet} & 299M & \multicolumn{2}{c|}{Manual} &  72.6 & 90.60 & \multicolumn{3}{c}{Manual}\\
\hline\hline
NASNET-A(4@1056)\cite{zoph2018learning} & 564M & - & - & 74.0 & 91.60 & False & 48000 & -\\
PNASNET\cite{liu2018progressive} & 588M & - & - &  74.2 &91.90 & False & 5400 & -\\
MnasNet-A1\cite{tan2019mnasnet} & 312M & - & - &  75.2 & 92.50 & False & $40000^{\S}$ & -\\
DARTS\cite{liu2018darts} & 574M & - & - & 73.3 & 91.30 & False & 24 & 288\\
SNAS\cite{xie2018snas} & 522M & - & - &  72.7 & 90.80 & False & 36 & 288\\
Proxyless-R (mobile)\cite{cai2018proxylessnas} & 320M & 62.6* & 84.7* & 74.6 & 92.20 & True & $300^{\ddagger}$ & $\geq$384\\
Single Path One-Shot\cite{guo2019single} & 319M & $68.7^{\dag}$ & - & 74.3 & - & True & 250 & 384 \\
Single Path One-Shot* & 323M & 68.2$\pm0.38$ & 88.28 & 74.3$\pm0.36$ & 91.79 & True & 250 & 384 \\
Random Search & $\leq$330M & $\leq$68.2 & $\leq$88.31 & $\leq$73.9 & $\leq$91.8 & True & 250 & 384 \\
\hline\hline
DSNAS & 324M & \textbf{74.4}$\bm{\pm0.22}$ & 91.54 & 74.3$\pm0.27$ & 91.90 & True & \multicolumn{2}{c}{420} \\
\hline
\end{tabular}
\caption{Results of choice block search. The time is measured based on NVIDIA TITAN X and accuracy is calculated on the validation set. * is our implementation with the original paper setting. $40000^{\S}$ is the GPU hour converted from 6912 TPUv2 hours with a ratio of roughly 1:6. $300^{\ddagger}$ is the GPU hour converted from V100 GPU with a ratio of 1:1.5. $68.7^{\dag}$ is the accuracy on the search set.}
\label{CB_result}
\end{table*}




\subsection{Time consumption breakup}
In last subsection, we show DSNAS can achieve comparable performance under the \textit{task-specific end-to-end} metric with much less computation than one-shot NAS methods. In this subsection, we further break up the time consumption of DSNAS into several specific parts, i.e. forward, backward, optimization and test\footnote{To clarify, we also do evaluation on testing set, retraining parameters is what we do not do.}, and conduct a controlled comparison with other differentiable NAS methods. We also hope such a detailed breakup can help readers gain insight into further optimizing our implementation. 

We first compare the computation time of SNAS and DSNAS on CIFAR-10 dataset. The average time of each splited part\footnote{The average time of each splited part in one batch is calculated on one NVIDIA TITAN X GPU with the same setting (batchsize 64) as in \cite{xie2018snas}.} is shown in Table \ref{tab:time_SNAS_DSNAS}. Under the same setting, our DSANS is almost \textbf{five} folds faster than SNAS and consumes only $\bm{1/n}$ of GPU memory as SNAS ($n$ is the total number of candidate operations in each edge). 
\begin{table}[H]
\centering
\begin{tabular}{c|c|c|c|c}
\hline
\multirow{2}*{Method} & \multicolumn{3}{c|}{Train} & \multirow{2}*{Test}\\
\cline{2-4}
 & Forward & Backward & Opt & \\
\hline\hline
SNAS & 0.26s  & 0.46s & 0.14s & 0.18s \\
DSNAS & 0.05s & 0.07s & 0.13s & 0.04s\\
\hline
\end{tabular}
\caption{Computation time of SNAS and DSNAS}\label{tab:time_SNAS_DSNAS}
\end{table}
We further compare the average time of each splited part between DSNAS and ProxylessNAS in a mobile setting on the ImageNet Classification task. As shown in Table \ref{tab:time_ProxylessNAS_DSNAS}, the average time \footnote{As shown in Table \ref{tab:complexity} that Proxyless NAS takes 2 times GPU memory as DSNAS, we use 8 TITAN X GPUs for ProxylessNAS and 4 for DSNAS to calculate the time.} is calculated on the same search space of ProxylessNAS \cite{cai2018proxylessnas} with a total batch size of 512. With a fair comparison, DSNAS is roughly \textbf{two} folds faster than ProxylessNAS.
\begin{table}[H]
\centering
\begin{tabular}{c|c|c|c|c}
\hline
\multirow{2}*{Method} & \multicolumn{3}{c|}{Train} & \multirow{2}*{Test}\\
\cline{2-4}
 & Forward & Backward & Opt & \\
\hline\hline
ProxylessNAS & 3.3s & 2.3s & 3.6s & 1.2s\\
DSNAS        & 1.9s & 1.3s & 2.6s & 0.9s\\
\hline
\end{tabular}
\caption{Computation time of ProxylessNAS and DSNAS}\label{tab:time_ProxylessNAS_DSNAS}
\end{table}


\section{Summary and future work}
In this work, we first define a \textit{task-specific end-to-end NAS} problem, under the evaluation metrics of which we scrutinize the efficiency of two-stage NAS methods. We then propose an efficient differentiable NAS framework, DSNAS, which optimizes architecture and parameters in the same round of back-propagation. Subnetworks derived from DSNAS are \textit{ready-to-deploy}. One competitive counterpart would be EfficientNet\cite{tan2019efficientnet}, which tries to bridge two stages with extra grid search on network scale after NAS. However, its total cost is larger than DSNAS. More accuracy gain can be achieved in DSNAS if scales are searched similarly. As a framework, DSNAS is orthogonal to the random wiring solution, which focuses on graph topology search \cite{xie2019exploring}. We look forward to their combination for a joint search of topology, operations, and parameters.


\section*{Acknowledgement}
This work is mainly done at SenseTime Research Hong Kong. SH and XL are also partially supported by by Hong Kong Research Grants Council General Research Fund No. 14200218 and Shun Hing Institute of Advanced Engineering Project No. MMT-p1-19.

{\small
\bibliographystyle{ieee_fullname}
\bibliography{main}
}
\appendix
\section{Detailed Settings of Experimental Results}
\textbf{Data Pre-processing    } We employ the commonly used pre-processing techniques in our experiments: A 224x224 crop is randomly sampled from an image or its horizontal flip, with a normalization on the pixel values per channel. 

\section{Details about the architectures}
\textbf{Structures of choice blocks \footnote{We follow the setting including choice blocks used in the released implementation of SPOS\cite{guo2019single}.}    }
    \begin{figure}[h!]
    \includegraphics[width=3.3in]{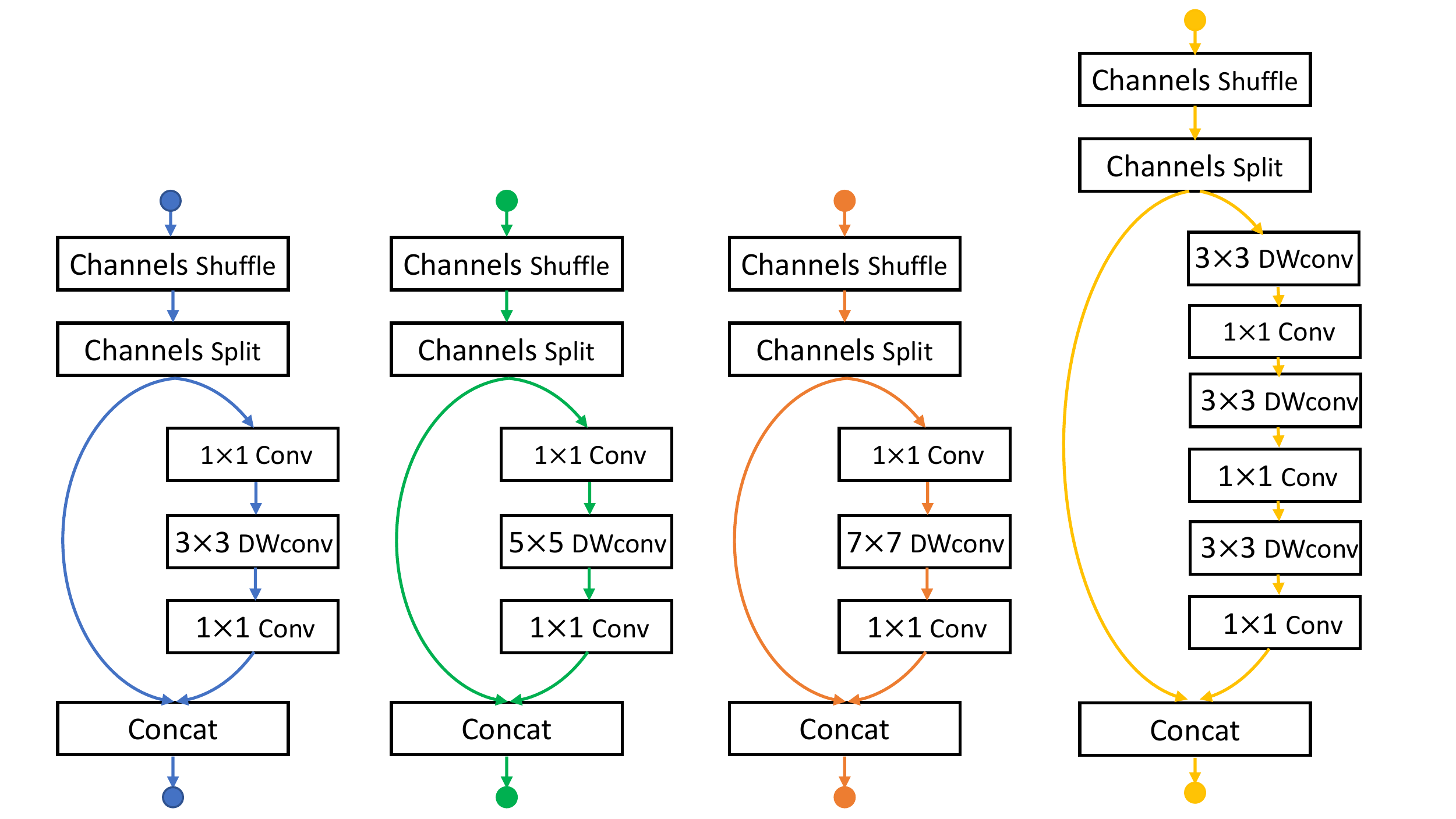}
    \caption{Choice blocks with stride=1. Choice blocks in search space. From left to right: Choice\_3, Choice\_5, Choice\_7, Choice\_x.}
    \end{figure}
    \begin{figure}[h!]
    \includegraphics[width=3.3in]{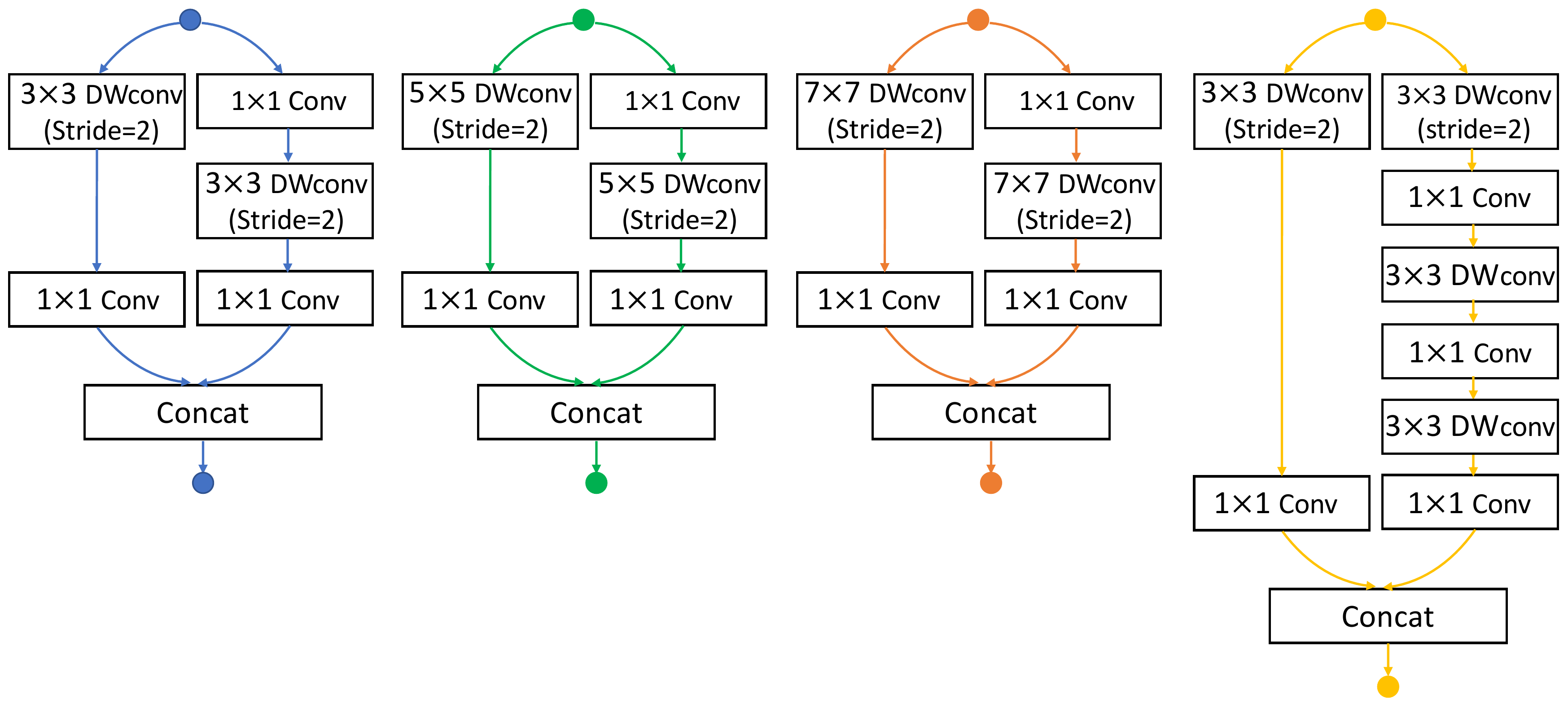}
    \caption{Choice blocks with stride=2. Choice blocks in search space. From left to right: Choice\_3, Choice\_5, Choice\_7, Choice\_x.}
    \end{figure}

\textbf{Supernet architecture    } 
\begin{table}[H]
\caption{Supernet architecture. CB - choice block. GAP - global average pooling. FC - fully connected layer. Each line describes a sequence of 1 or more identical layers, repeated \textit{Repeat} times. All layers in the same sequence have the same number of output channels. The first layer of each sequence has a stride \textit{Stride} and all others use stride 1.}\label{supernet_arch}
\centering
\begin{tabular}{c|c|c|c|c}
\hline\hline
Input & Block & Channels & Repeat  & Stride \\
\hline\hline
$224^2\times3$ & $3\times3$ Conv & 16 & 1 & 2\\
$112^2\times16$ & CB & 64 & 4 & 2\\
$56^2\times64$ & CB & 160 & 4 & 2\\
$28^2\times160$ & CB & 320 & 8 & 2\\
$14^2\times320$ & CB & 640 & 4 & 2\\
$7^2\times640$ & $1\times1$ Conv & 1024 & 1 & 1\\
$7^2\times1024$ & GAP & - & 1 & -\\
1024 & FC & 1000 & 1 & -\\
\hline\hline
\end{tabular}
\end{table}
\textbf{Structures of searched architectures    }
\begin{figure}[H]
    \centering
    \includegraphics[width=3.3in]{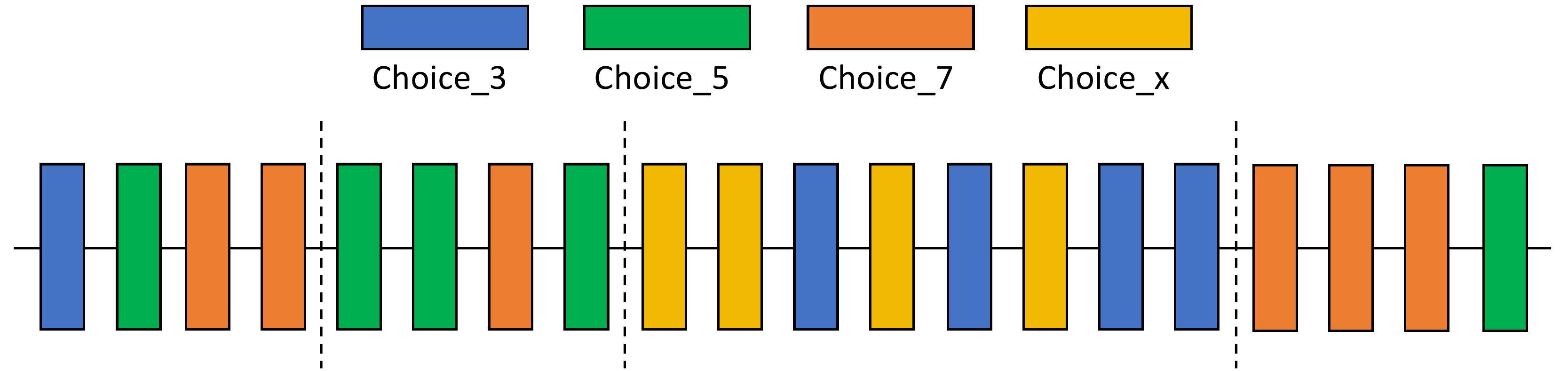}
    \caption{Our implementation of building block search result in SPOS\cite{guo2019single}.}
    \label{fig: SPOS_search_arch}
\end{figure}
\begin{figure}[H]
    \vspace{-4.8mm}
    \centering
    \includegraphics[width=3.3in]{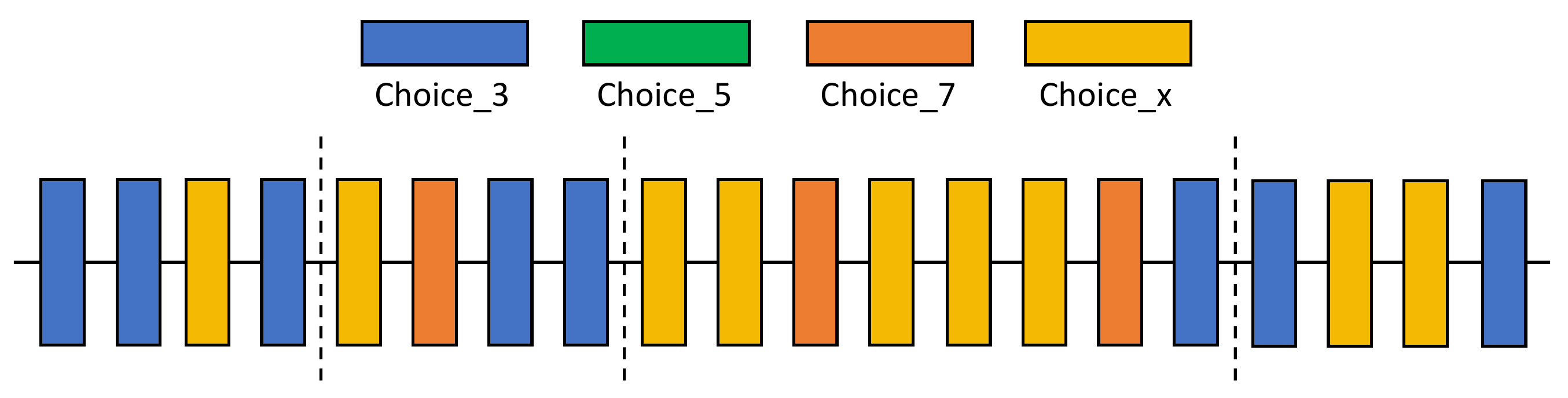}
    \caption{Building block search result in Sec. 4. based on one-stage \textit{searching} process, i.e., \texttt{DSANSsearch240}.}
    \label{fig: search_arch}
\end{figure}
\end{document}